\documentclass{article} 
\usepackage{iclr2027_conference,times}

\usepackage{amsmath,amsfonts,bm}

\def\eqref#1{equation~\ref{#1}}

\def\1{\bm{1}}

\DeclareMathAlphabet{\mathsfit}{\encodingdefault}{\sfdefault}{m}{sl}
\SetMathAlphabet{\mathsfit}{bold}{\encodingdefault}{\sfdefault}{bx}{n}

\usepackage{hyperref}
\usepackage{url}
\usepackage{amsmath,amssymb,booktabs,graphicx,microtype}
\usepackage{algorithm}
\usepackage{enumitem}
\usepackage[noend]{algpseudocode}

\title{AttnSR: Attention Step Reward for Long-Horizon Agentic RL}

\author{%
\begin{minipage}[t]{0.97\textwidth}
\centering\normalfont
\mbox{\textbf{Mengsong Wu}\textsuperscript{1,2,3\textdagger}}\quad
\mbox{\textbf{Tianchen Zhou}\textsuperscript{3,4\textdagger}}\quad
\mbox{\textbf{Zhaochen Su}\textsuperscript{5\textdagger}}\quad
\mbox{\textbf{Zhenhua Liu}\textsuperscript{3\textdaggerdbl}}\\[3pt]
\mbox{\textbf{Zhe Cui}\textsuperscript{3*}}\quad
\mbox{\textbf{Aimin Zhou}\textsuperscript{1,2*}}\endgraf\vspace{6pt}
{\footnotesize
\textsuperscript{1}Shanghai Institute of AI for Education, East China Normal University\\
\textsuperscript{2}Shanghai Innovation Institute\quad
\textsuperscript{3}Hithink Research\\
\textsuperscript{4}School of Data Science and Engineering, East China Normal University\\
\textsuperscript{5}The Hong Kong University of Science and Technology\endgraf}
\vspace{3pt}
{\footnotesize\texttt{cuizhe@myhexin.com}\quad
\texttt{amzhou@cs.ecnu.edu.cn}\endgraf}
\end{minipage}%
}

\newcommand{\method}{\textsc{AttnSR}}
\newcommand{\alf}{\textsc{ALFWorld}}
\newcommand{\webshop}{\textsc{WebShop}}
\newcommand{\science}{\textsc{ScienceWorld}}

\iclrfinalcopy 
\begin{document}
\maketitle
\lhead{Manuscript}
\setlength{\skip\footins}{7pt plus 1pt minus 1pt}
\setlength{\footnotesep}{5.5pt}
\begingroup
\renewcommand{\thefootnote}{\fnsymbol{footnote}}
\footnotetext[2]{Equal Contributions\quad\textsuperscript{\textdaggerdbl}Project Lead\quad\textsuperscript{*}Corresponding authors}
\endgroup

\begin{abstract}
Long-horizon LLM agents make multi-step decisions, yet outcome-reward-only reinforcement learning provides little direct supervision for individual steps. We introduce \method{}, a step-level reward augmentation method that derives reward allocations from policy attention extracted during training replay, without a separately trained reward model or additional rollouts. \method{} traces backward through the policy model attention graph to estimate the relevance of earlier steps to the final response. The terminal step, measured relative to a running baseline, determines whether these rewards reinforce or penalize the selected steps. In a controlled Qwen2.5-7B-Instruct study, \method{} consistently outperforms the PPO baseline across all evaluated checkpoints on \alf{}, raising final held-out success from 91.2\% to 97.1\%. The ablation experiment shows that the benefit depends on how intermediate rewards are assigned, rather than simple accumulation of step rewards over time. \method{} also improves \science{} at every checkpoint while cutting environment interactions by 29.9\%. These results support policy attention as a practical prior for step-level reward allocation in long-horizon agentic RL. Code is available at \url{https://github.com/fairyshine/AttnSR}.

\end{abstract}
\section{Introduction}
Reinforcement learning (RL) with verifiable rewards~\citep{shao2024deepseekmath,deepseek2025r1} is increasingly applied to agents that interleave reasoning and tool use~\citep{putta2024agentq,wang2025ragen,yao2023react}. In environments such as \alf{}, \webshop{}, and \science{}, agents make sequences of decisions to complete a task~\citep{shridhar2021alfworld,yao2022webshop,wang2022scienceworld}. 
Vanilla PPO uses the outcome reward to estimate value targets and policy advantages~\citep{schulman2017ppo,schulman2016gae}, but says little about how intermediate actions contribute.
Assigning credit to earlier decisions thus remains a core challenge in long-horizon RL~\citep{zhang2026creditsurvey}.

Step-level rewards address this challenge by providing more localized training signals, but obtaining them can require step annotations, additional environment interactions, or a separately trained evaluator~\citep{lightman2023verify,setlur2024rewarding,kazemnejad2025vineppo}. This motivates using signals already available within the policy model to guide intermediate rewards, without incurring the substantial training and environment-interaction costs.

Attention patterns in the policy model may provide such a signal, since they record how later decisions attend to earlier actions and observations. We hypothesize that these relationships provide a useful prior for allocating intermediate rewards, although attention alone does not establish causal contributions~\citep{jain2019attention,abnar2020quantifying}. The signal can be extracted during the existing forward pass of the policy model over collected trajectories. The method adds little computation and requires neither a separate attribution backward pass nor a learned progress estimator.

Building on this hypothesis, we introduce Attention Step Reward (\method), which uses the attention of the policy model to identify relevant earlier steps and turns the task outcome into intermediate training signals. Starting from the final step, it traces attention links backward to estimate the relevance of earlier actions and observations. These estimates determine which intermediate steps receive reward, while weaker or less consistent attention reduces their overall magnitude. Outcomes above the running average score yield positive intermediate rewards, and those below it yield negative rewards. These signals can in principle be used directly by step-level RL algorithms or placed at assistant-step endpoints for token-level RL. We study the latter integration in PPO, where the added rewards change value targets and policy advantages while retaining the critic architecture and losses.

We evaluate \method{} on \alf{}, \science{}, and \webshop{} using Qwen2.5-7B-Instruct with matched token-level PPO training. ALFWorld success rises from 91.2\% to 97.1\%, with larger gains on unseen tasks, whereas the positional ablation reaches 92.3\%. On ScienceWorld, \method{} yields higher official scores at all four reported checkpoints and 29.9\% fewer mean environment steps. These results suggest that policy attention can serve as a useful signal for intermediate reward allocation, offering a way to enrich outcome-based supervision without additional rollouts or a separately trained reward model.

Our contributions are summarized below.
\begin{itemize}[leftmargin=*]
\setlength{\itemsep}{0pt}
    \item We hypothesize that policy attention provides a relevance prior over interaction steps by indicating how later decisions attend to earlier actions and observations. Based on this hypothesis, we use attention-derived dependencies to distribute outcome information across steps for temporal credit assignment, treating attention as a heuristic relevance signal, not evidence of causal contribution.
    \item We introduce \method, a step-level reward augmentation scheme that combines attention-based allocation, confidence gating, and outcome centering. The signals can be used directly as step-level rewards or aligned with assistant-step endpoints for token-level objectives, and we study the latter integration with PPO. It needs neither counterfactual rollouts nor a separately trained reward model.
    \item We evaluate \method{} across three agentic benchmarks with ablation and trajectory analysis to characterize its performance gains and potential boundary conditions across tasks with different interaction lengths. The results are consistent with a stronger benefit on complex, long-horizon tasks, while suggesting more limited gains when trajectories are short.
\end{itemize}

\section{Problem Formulation}
\label{sec:problem}
We model each LLM-based agent interaction trajectory as a multi-turn conversation. Episode $i$ starts from a task query and contains $K_i$ decision steps. Each step $S_{i,k}$ comprises an assistant message $A_{i,k}$ and its environment observation $O_{i,k}$ when present, and the final message may have no observation. Let $\mathcal{I}_{i,k}$ be the assistant token indices at step $k$, and $e_{i,k}\in\mathcal{I}_{i,k}$ its last trainable token, the \emph{step endpoint}. We supply PPO with a task outcome reward $R_i\in[0,1]$. In the canonical runs, $R_i$ is binary success for \alf{}, the normalized official score for \webshop{}, and the normalized final score for \science{}. Format-error trajectories receive $R_i=0$. Thus the binary success metric for \webshop{} and \science{} is an evaluation threshold on the official score, not a replacement for the training reward. Table~\ref{tab:notation} in Appendix~\ref{app:notation} summarizes the notation.

Our terminal-only PPO baseline aligns rewards with trajectory tokens, indexed by $u$, placing the outcome at the final assistant endpoint as
\begin{equation}
r_{i,u}^{\mathrm{PPO}}=R_i\,\mathbb{1}[u=e_{i,K_i}],
\label{eq:terminal-reward}
\end{equation}
where $\mathbb{1}[\cdot]$ is the indicator function. Observation tokens provide context but are masked from actor and critic losses. TD residuals and GAE compute value-function targets and policy advantages over valid assistant tokens, skipping masked positions~\citep{schulman2016gae,schulman2017ppo}. The outcome reward evaluates the whole trajectory without specifying contributions of individual actions.

Let $s_{i,k}$ denote a signed shaping reward assigned to intermediate step $k<K_i$. \method{} augments the baseline reward sequence as
\begin{equation}
r_{i,u}^{\mathrm{AttnSR}}=r_{i,u}^{\mathrm{PPO}}+
\sum_{k=1}^{K_i-1}s_{i,k}\,\mathbb{1}[u=e_{i,k}].
\label{eq:reward-augmentation}
\end{equation}
This retains $R_i$ at the final endpoint and adds $s_{i,k}$ at intermediate assistant endpoints. Algorithms that consume step-level rewards can use these signals directly in principle, and Equation~\ref{eq:reward-augmentation} specifies their token-level placement for the PPO implementation studied here. Our goal is to derive $s_{i,k}$ from actor attention and terminal outcomes. Section~\ref{sec:method} defines its allocation, confidence gate, and sign.

\section{\method: Actor Attention for Step Credit}
\label{sec:method}
\method{} constructs intermediate rewards by tracing how later decisions attend to earlier interaction steps. It uses these relationships to allocate rewards, with their magnitude and sign determined by the reliability of the attention signal and the trajectory outcome relative to a running baseline.

\subsection{Where to assign credit: a step dependency graph}
For each selected transformer layer $l$, we construct a directed graph over interaction steps. Its components $G_{i,\mathrm{AA}}^{(l)}[j,k]$ and $G_{i,\mathrm{AO}}^{(l)}[j,k]$ measure attention from the assistant message at step $j$ to the assistant and observation messages at step $k$, respectively. We sum causal attention probabilities over each historical message key tokens, then average over step $j$ query tokens and heads. Only $k<j$ edges are retained, and missing observations contribute zero AO mass. The combined graph is $G_i^{(l)}=0.5G_{i,\mathrm{AA}}^{(l)}+0.5G_{i,\mathrm{AO}}^{(l)}$.

\begin{figure}[t]
	\begin{center}
		\includegraphics[width=\linewidth]{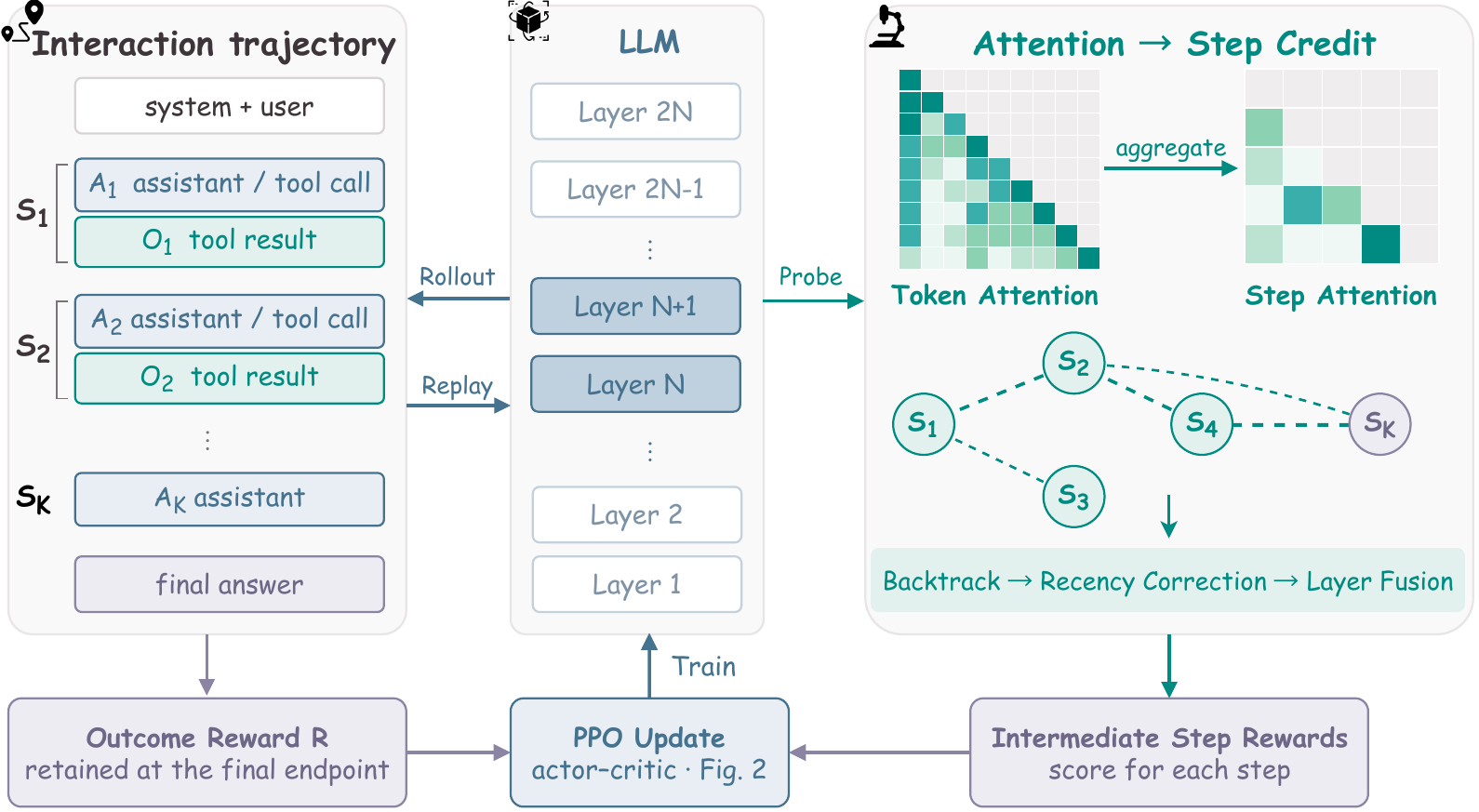}
	\end{center}
	\caption{Overview of \method. Actor replay supplies attention for a step graph. Backtracking from the terminal step, recency correction, and layer fusion yield an allocation, and confidence and the centered outcome scale it into signed intermediate rewards. The outcome reward is retained. Graph edges link later steps to earlier dependencies, and episode indices are omitted.}
	\label{fig:method}
\end{figure}

During actor replay for old log-probabilities, we capture Q/K after rotary position embedding~\citep{su2024roformer} and before fused attention~\citep{dao2022flashattention}, reconstructing the packed tensor/context-parallel sequence. Query blocks use full causal-prefix normalization before message aggregation, and the historical-step submatrix is not renormalized, which preserves its mass relative to other context tokens. Figure~\ref{fig:method} illustrates how token attention becomes a step graph for reward construction.

Relevance then flows backward from the terminal step, independently within each layer. Writing $a_{i,k}^{(l)}\ge 0$ for the relevance layer $l$ assigns to step $k$,
\begin{equation}
a_{i,K_i}^{(l)}=1, \qquad
a_{i,k}^{(l)}=\rho\sum_{j>k} a_{i,j}^{(l)}G_i^{(l)}[j,k],
\label{eq:backtrack}
\end{equation}
where $j\le K_i$. Each backward path contributes its edge-weight product with decay $\rho=0.85$ per hop, capturing direct and indirect dependencies. This graph-hop decay differs from the temporal-distance correction below. Only $k<K_i$ scores enter reward allocation.

\subsection{How much to trust attention: correction and consensus}
To reduce recency preference, define terminal distance $\Delta_{i,k}=K_i-k$ and proximity $d_{i,k}=\exp(-(\Delta_{i,k}-1)/\kappa)$. For $K_i>1$, we standardize log relevance and proximity over intermediate steps as
\begin{equation}
z_{i,k}^{(l)}=\operatorname{norm}\!\left(\log\max(a_{i,k}^{(l)},\epsilon)\right)+\eta\operatorname{norm}(d_{i,k}),\qquad
p_{i,k}^{(l)}=\frac{\exp(z_{i,k}^{(l)}/T)}{\sum_{h<K_i}\exp(z_{i,h}^{(l)}/T)}.
\label{eq:bias}
\end{equation}
Here $\epsilon=10^{-12}$ and $\operatorname{norm}(x)=(x-\operatorname{mean}(x))/\operatorname{std}(x)$ uses population standard deviation, returning zero for a single element or standard deviation at most $10^{-6}$. We use $T=1$, $\kappa=4$, and $\eta=-0.25$. The negative proximity term reduces allocation to recent steps but does not set the sign.

Because attribution varies across layers~\citep{abnar2020quantifying}, we fuse their independently computed allocations by a geometric mean,
\begin{equation}
q_{i,k}=\exp\left(\frac{1}{L}\sum_{l\in\mathcal L}\log\max\left(p_{i,k}^{(l)},\epsilon\right)\right),
\qquad
p_{i,k}=\frac{q_{i,k}}{\sum_{h<K_i}q_{i,h}},
\label{eq:consensus}
\end{equation}
where $L=|\mathcal L|$ is the number of selected layers. Geometric fusion favors consistent support across layers, and the normalization runs over intermediate steps.

Confidence controls the strength of the attention-derived reward. For layer $l$, define the terminal covered mass as
\begin{equation}
m_i^{(l)}=\sum_{k<K_i}\left(G_{i,\mathrm{AA}}^{(l)}[K_i,k]+G_{i,\mathrm{AO}}^{(l)}[K_i,k]\right)\in[0,1].
\label{eq:covered-mass}
\end{equation}
This uses the summed AA and AO masses before the equal-weight graph mixture. For $K_i>2$, define the normalized entropy
\begin{equation}
H_i^{(l)}=-\frac{\sum_{k<K_i}p_{i,k}^{(l)}\log\max(p_{i,k}^{(l)},\epsilon)}{\log(K_i-1)}.
\label{eq:allocation-entropy}
\end{equation}
For $K_i=2$, the unique intermediate step has allocation one and we set $H_i^{(l)}=0$. The confidence components are
\begin{align}
C_{i,\mathrm{mass}}^{(l)}&=\operatorname{clip}(m_i^{(l)}/0.20,0,1),\\
C_{i,\mathrm{focus}}^{(l)}&=\operatorname{clip}((1-H_i^{(l)})/0.15,0,1),\\
C_{i,\mathrm{layer}}&=1-\operatorname{mean}_l\operatorname{TV}(p_i^{(l)},p_i),\\
C_i&=\operatorname{mean}_l\!\left(C_{i,\mathrm{mass}}^{(l)}C_{i,\mathrm{focus}}^{(l)}\right)C_{i,\mathrm{layer}}\in[0,1],
\label{eq:confidence}
\end{align}
where $\operatorname{clip}(x,0,1)$ saturates to $[0,1]$ and $\operatorname{TV}(p,q)=\frac12\sum_k|p_k-q_k|$. Covered mass, concentration, and layer agreement measure historical attention, selectivity, and consistency, respectively. The values $0.20$ and $0.15$ are saturation thresholds. The resulting confidence is a shaping gate, not a calibrated probability of correct attribution.

Zero total intermediate relevance sets that layer focus confidence to zero. For $K_i=1$, we skip allocation, set $C_i=0$, and retain terminal-only rewards.

\begin{algorithm}[htbp]
	\caption{\method{} reward construction for one batch}
	\label{alg:attnsr}
	\begin{algorithmic}[1]
		\Require completed trajectory with endpoints $e_{i,k}$ and rewards $R_i$, selected layers $\mathcal L$, and the previous baseline if initialized
		\Ensure token rewards $r_{i,u}$ and updated baseline $\bar R_t$
		\State capture detached post-RoPE $Q,K$ during actor replay for old log-probabilities
		\State initialize $r_{i,u}\gets R_i\mathbb{1}[u=e_{i,K_i}]$ and $C_i\gets0$ for each turn
		\State compute batch mean $M_t$, then initialize or update $\bar R_t$ by Equation~\ref{eq:budget}
		\For{each episode $i$ with $K_i>1$}
		\For{each layer $l\in\mathcal L$}
		\State aggregate causal attention into AA/AO step graphs and set $G_i^{(l)}\gets\tfrac12(G_{i,\mathrm{AA}}^{(l)}+G_{i,\mathrm{AO}}^{(l)})$
		\State backtrack relevance and apply recency correction to obtain $p_i^{(l)}$ \Comment{Eqs.~\ref{eq:backtrack}--\ref{eq:bias}}
		\State compute mass and focus confidence, setting focus to zero if intermediate relevance is zero
		\EndFor
		\State fuse layer allocations into $p_i$ and compute $C_i$ \Comment{Eqs.~\ref{eq:consensus}--\ref{eq:confidence}}
		\State add $\beta C_i(R_i-\bar R_t)p_{i,k}$ at $u=e_{i,k}$ for $k<K_i$ \Comment{retain outcome reward}
		\EndFor
		\State \Return $r_{i,u}$, $\bar R_t$
	\end{algorithmic}
\end{algorithm}

\subsection{Whether to reinforce a step: a centered budget}
Attention indicates computational relevance, not whether a step contributed positively to success. We determine reward sign relative to a running outcome baseline. For collected training batch $t$ with mean outcome reward $M_t$, we initialize or update the baseline before shaping as
\begin{equation}
\bar R_t=\begin{cases}
M_1, & t=1,\\
0.99\bar R_{t-1}+0.01M_t, & t>1,
\end{cases}
\qquad B_i=\beta\,C_i(R_i-\bar R_t),
\label{eq:budget}
\end{equation}
where $\beta=0.25$. The detached baseline is shared across the batch. Outcomes above it yield positive budgets, and those below it yield negative budgets. Thus sign reflects relative outcome quality, not binary success. For $k<K_i$, we assign
\begin{equation}
s_{i,k}=B_i\,p_{i,k}=\beta\,C_i(R_i-\bar R_t)\,p_{i,k},
\label{eq:step-reward}
\end{equation}
at endpoint $e_{i,k}$, retaining $R_i$ at $e_{i,K_i}$. For $K_i>1$, intermediate rewards sum to $B_i$, and zero confidence yields no shaping. Format-error trajectories retain environment reward $R_i=0$ and participate in the EMA update. Algorithm~\ref{alg:attnsr} summarizes reward construction for one batch.

Since $C_i,R_i,\bar R_t\in[0,1]$, $\sum_{k<K_i}|s_{i,k}|=|B_i|\leq\beta$, so the upper bound is length-independent, while the realized budget depends on confidence and outcome. The undiscounted shaped return is $R_i+B_i$, so policy invariance is not guaranteed. This is a training heuristic distinct from return-preserving redistribution~\citep{arjona-medina2019rudder}, and evaluation uses the original task metrics.

\subsection{Complexity and integration}
Blockwise aggregation into step buckets avoids storing full token attention matrices but retains quadratic query--key arithmetic and selected-layer buffers. It reuses the actor forward pass without additional rollouts or a separately trained reward model. Wall-clock and peak-memory overheads remain unmeasured. Rewards enter PPO before advantage estimation, keeping its critic and losses.

Actor attention is detached when constructing rewards, while policy probabilities remain differentiable during optimization. Given pre-update critic values $V_u^{\mathrm{old}}$ and unnormalized GAE advantages $\hat A_u$, the canonical PPO training signals are
\begin{equation}
\widetilde A_u=\operatorname{sg}\!\left(\operatorname{Norm}(\hat A)_u\right),
\qquad
y_u=\operatorname{sg}\!\left(\hat A_u+V_u^{\mathrm{old}}\right),
\label{eq:actor-critic-targets}
\end{equation}
where $\operatorname{Norm}$ uses the response mask and $\operatorname{sg}$ denotes stop-gradient. The actor uses $\widetilde A_u$ in its clipped surrogate, and the critic fits returns $y_u$, constructed \emph{before} advantage normalization, using the existing clipped value loss with reference $V^{\mathrm{old}}$. Both signals remain fixed within the batch, and neither loss backpropagates through GAE or reward construction.

\section{Experiments}
\label{sec:experiments}

\subsection{Experimental Setup}
\label{sec:experimental-setup}

\paragraph{Agentic Tasks.} We evaluate on \alf{} (274 held-out tasks, 140 seen and 134 unseen), \webshop{} (500 test tasks), and \science{} (1,819 test tasks)~\citep{shridhar2021alfworld,yao2022webshop,wang2022scienceworld}, covering household interaction, online shopping, and scientific procedures. ALFWorld official score equals task success rate. WebShop and ScienceWorld additionally provide graded scores for partial progress, reported separately from binary success. Their evaluators mark success at normalized scores of at least $0.99$ and $0.999$, respectively.

\begin{figure}[t]
	\begin{center}
		\includegraphics[width=\linewidth]{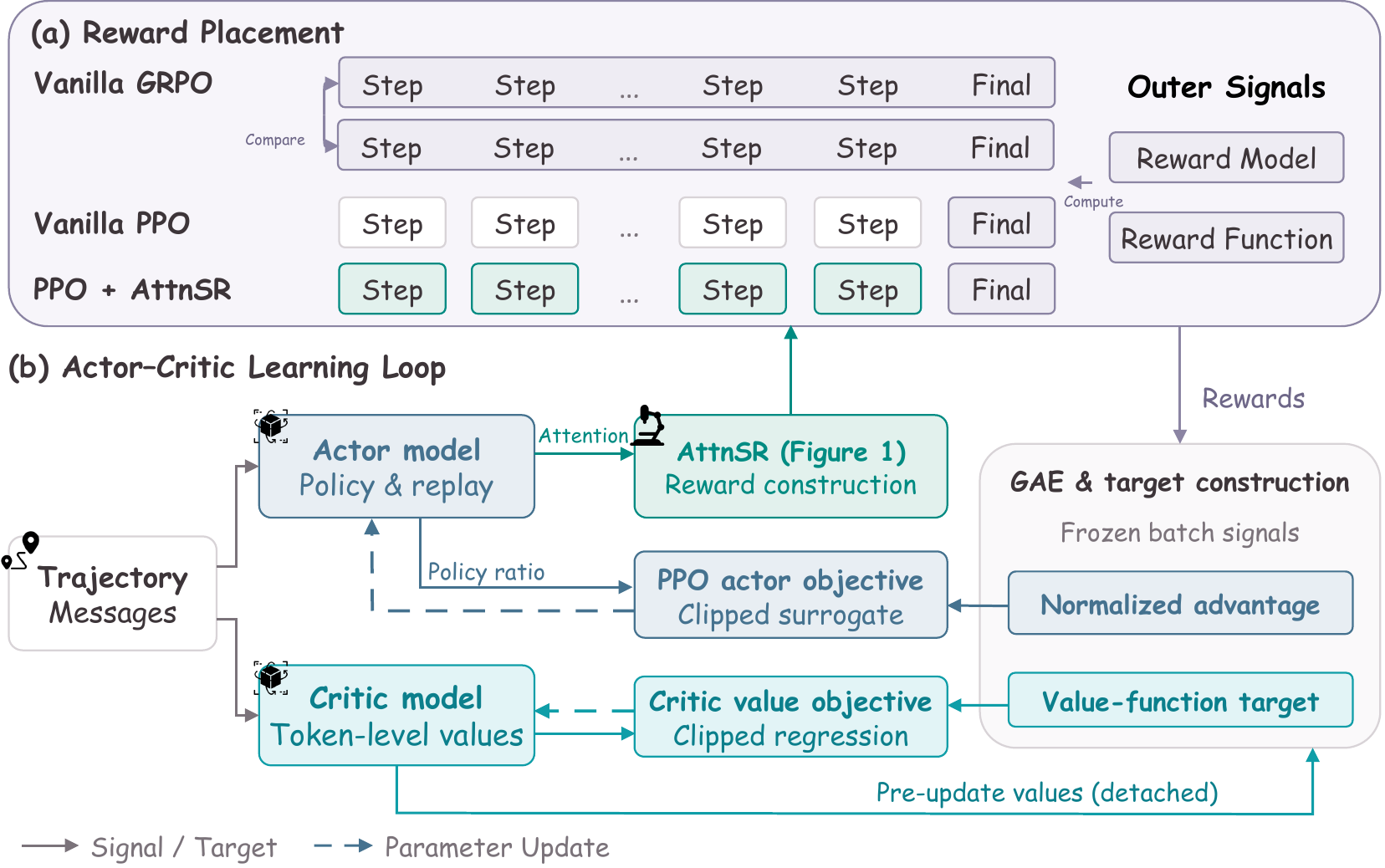}
	\end{center}
	\caption{Integration of \method{} into actor--critic RL. (a)~The PPO integration of \method{} adds intermediate rewards while retaining the outcome reward. (b)~Detached actor attention supplies rewards, and pre-update critic values support GAE and target construction. Normalized advantages guide the actor, while estimated returns supervise the critic as in Equation~\ref{eq:actor-critic-targets}. Targets remain fixed during optimization. Solid arrows carry signals or targets, and dashed arrows denote updates.}
	\label{fig:actor-critic}
\end{figure}

\paragraph{RL Training.} We use Qwen2.5-7B-Instruct~\citep{qwen2024qwen25} with a frozen base model and rank-64 LoRA adapters~\citep{hu2022lora} for actor and critic. Both learning rates are $5\times10^{-6}$, the global batch size is 8, and the reference KL coefficient is 0.02. PPO uses one rollout per prompt and one optimization epoch per batch. Separate policies train on each benchmark training split, excluding held-out tasks. Table~\ref{tab:training-size} in Appendix~\ref{app:implementation} reports data sizes and update counts. Figure~\ref{fig:actor-critic} shows how detached actor attention changes policy advantages and value targets through intermediate rewards, retaining the PPO losses. We evaluate after each of four data epochs. Training uses VeRL~\citep{sheng2025verl}, Megatron~\citep{shoeybi2019megatron}, and vLLM~\citep{kwon2023vllm} on 8$\times$H200 GPUs (140\,GB each), with a 32,768-token context and no additional per-turn token or turn-count cap. Attention extraction uses $L=6$ layers and 128-token query blocks. Appendix~\ref{app:implementation} lists all hyperparameters.

\paragraph{Baselines and Controls.} Terminal-only PPO matches \method{} in model, rollouts, optimizer, and evaluation.
Linear Step Reward, described in Section~\ref{sec:ablation}, controls for positional shaping on ALFWorld.
GRPO uses four rollouts per prompt for one epoch.
Alongside the 7B GRPO reference, we compare 3B GRPO with an AttnSR token-advantage variant, as detailed in Appendix~\ref{app:3b-grpo}.
PPO/GRPO optimization budgets differ.
We additionally compare adaptations of TARL~\citep{tan2025tarl}, AgentPRM~\citep{xi2025agentprm}, and FlowTracer~\citep{dong2026flowtracer} on ALFWorld after four epochs of Qwen2.5-7B-Instruct LoRA training, using their final checkpoints. Appendix~\ref{app:additional-baselines} gives the details.

\paragraph{Evaluation Protocol.} We decode greedily once per held-out task and report full-set metrics. The WebShop validation split is reserved for training-time validation. The main comparison uses epoch 4 without selecting the best test-set checkpoint. Format errors violate the required tool-call protocol and receive outcome reward zero without an additional training penalty. For statistical comparisons, exact-input matching retains all 274 ALFWorld and 500 WebShop tasks and 1,488 of 1,819 ScienceWorld tasks, as described in Appendix~\ref{app:trajectory-diagnostics}. A two-sided exact McNemar test~\citep{mcnemar1947} assesses binary success differences, and a paired bootstrap~\citep{efron1979bootstrap} with 10,000 resamples estimates 95\% percentile intervals for success-rate and official-score differences. These intervals quantify sampling uncertainty over the matched held-out tasks, conditional on the trained policies.

\begin{table}[htbp]
\caption{Task performance with Qwen2.5-7B-Instruct. (a)~Main Result across three benchmarks. (b)~ALFWorld success. All values are percentages (\%).}
\label{tab:method-comparison}
\centering
\small
\setlength{\tabcolsep}{3pt}
\renewcommand{\arraystretch}{1.15}
\newsavebox{\MainResultBox}
\newsavebox{\AlfworldMethodBox}
\newlength{\MainResultRowPad}
\setlength{\MainResultRowPad}{0pt}
\newcommand{\MainResultStrut}{\rule[\dimexpr-1.15\dp\strutbox-\MainResultRowPad\relax]{0pt}{\dimexpr1.15\ht\strutbox+1.15\dp\strutbox+2\MainResultRowPad\relax}}
\newcommand{\MainResultTable}{%
\begin{tabular*}{0.72\linewidth}{@{\extracolsep{\fill}}lrrrrrrr@{}}
\toprule
\MainResultStrut & \multicolumn{3}{c}{\alf{}} & \multicolumn{2}{c}{\webshop{}} & \multicolumn{2}{c}{\science{}}\\
\cmidrule(lr){2-4}\cmidrule(lr){5-6}\cmidrule(l){7-8}
\MainResultStrut Method & Score & Seen & Unseen & Score & Success & Score & Success\\
\midrule
\MainResultStrut GRPO$^{\dagger}$ & 69.71 & 75.00 & 64.18 & 67.21 & 37.80 & 30.19 & 16.66\\
\MainResultStrut PPO & 91.24 & 92.86 & 89.55 & \textbf{70.74} & 40.20 & 44.53 & 28.59\\
\MainResultStrut PPO + AttnSR & \textbf{97.08} & \textbf{96.43} & \textbf{97.76} & 69.71 & \textbf{41.00} & \textbf{46.35} & \textbf{30.90}\\
\bottomrule
\end{tabular*}}
\sbox{\MainResultBox}{\MainResultTable}
\sbox{\AlfworldMethodBox}{%
\fontsize{8}{9}\selectfont%
\renewcommand{\arraystretch}{1.05}%
\begin{tabular*}{0.25\linewidth}{@{\extracolsep{\fill}}lr@{}}
\toprule
Method & Success\\
\midrule
Raw & 26.64\\
GRPO$^{\dagger}$ & 69.71\\
PPO & 91.24\\
TARL & 96.72\\
AgentPRM & 91.97\\
FlowTracer & 95.62\\
PPO + AttnSR & \textbf{97.08}\\
\bottomrule
\end{tabular*}}
\setlength{\MainResultRowPad}{\dimexpr(\ht\AlfworldMethodBox+\dp\AlfworldMethodBox-\ht\MainResultBox-\dp\MainResultBox)/10\relax}
\sbox{\MainResultBox}{\MainResultTable}
\begin{minipage}[t]{0.72\linewidth}
\centering
\textbf{(a) Main Result}\par\smallskip
\usebox{\MainResultBox}
\end{minipage}\hfill
\begin{minipage}[t]{0.25\linewidth}
\centering
\textbf{(b) ALFWorld}\par\smallskip
\raisebox{\dimexpr\dp\AlfworldMethodBox-\dp\MainResultBox\relax}{\usebox{\AlfworldMethodBox}}
\end{minipage}
\par\smallskip
\begin{minipage}{\linewidth}
\footnotesize Raw is the instruction-tuned model before RL. $^{\dagger}$GRPO uses epoch 1 with $n=4$, and all other trained policies use epoch 4, with $n=1$ for PPO and AttnSR. Optimization budgets differ. ALFWorld score equals success, while other tasks report score and success separately. Panel (b) reuses the canonical GRPO, PPO, and AttnSR results, and Appendix~\ref{app:additional-baselines} details the other baselines.
\end{minipage}
\end{table}

\subsection{Main Results}
\label{sec:results}
As shown in Table~\ref{tab:method-comparison}, at epoch 4 \method{} shows the clearest benefit on ALFWorld, smaller numerical gains on ScienceWorld, and mixed outcomes on WebShop. ALFWorld success rises from 91.24\% to 97.08\% (266/274), reducing failures from 24 to 8. ScienceWorld official score increases from 44.53\% to 46.35\%, and success from 28.59\% to 30.90\%. WebShop success rises from 40.20\% to 41.00\%, but official score falls from 70.74\% to 69.71\%.

\paragraph{Additional ALFWorld Baselines.} Table~\ref{tab:method-comparison}(b) compares final success with TARL (96.72\%), AgentPRM (91.97\%), and FlowTracer (95.62\%). \method{} attains the highest observed success, 97.08\%, although its margin over TARL is only one task (266 versus 265 of 274). These single-run results are descriptive, and the paired tests below concern the canonical PPO/AttnSR comparison.

\paragraph{Statistical Evidence.} Tests use exact-input-matched tasks rather than the full-set estimates in Table~\ref{tab:method-comparison}. On ALFWorld, 20 tasks are solved only by \method{} and 4 only by PPO, giving $p=0.002$ and a 95\% gain interval of $[+2.55,+9.49]$ points. Only ALFWorld remains significant after Bonferroni correction across the three success tests. ScienceWorld gains $+2.49$ points on 1,488 matched tasks, versus $+2.31$ on the full set, but its unadjusted $p=0.040$ does not survive correction. As reported in Appendix~\ref{app:significance}, its score interval and both WebShop intervals include zero.

\subsection{Ablation, Control, and Transfer Analysis}
\label{sec:ablation}
\paragraph{Positional Reward Control.} Linear Step Reward tests whether a simple temporal allocation can reproduce the ALFWorld gain. It retains the outcome reward, outcome centering, and budget scale $0.25$, but replaces attention allocation with increasing temporal weights and removes confidence gating. For $K_i>1$,
\begin{equation}
w^{\mathrm{lin}}_{i,k}=\frac{k}{\sum_{j=1}^{K_i-1}j},\qquad
s^{\mathrm{lin}}_{i,k}=0.25(R_i-\bar R_t^{\mathrm{lin}})w^{\mathrm{lin}}_{i,k}.
\label{eq:linear}
\end{equation}
Each run maintains its own outcome baseline. The control reaches 92.34\% success, gaining 1.10 points over PPO versus 5.84 points for \method{}, as shown in Table~\ref{tab:linear}. The control thus recovers less than one fifth of the gain.

\begin{table}[!htbp]
\begin{minipage}[t]{0.42\linewidth}
\caption{ALFWorld success (\%). $\Delta$ is the gain over each block baseline in percentage points.}
\label{tab:linear}
\label{tab:qwen25-3b}
\centering
\scriptsize
\setlength{\tabcolsep}{4pt}
\renewcommand{\arraystretch}{0.87}
\begin{tabular}{lrr}
\toprule
Method & Succ. (\%) & $\Delta$ (pp)\\
\midrule
PPO & 91.24 & --\\
$+$Linear Step Reward & 92.34 & +1.10\\
$+$AttnSR & \textbf{97.08} & \textbf{+5.84}\\
\midrule
PPO \textit{Asynchronous} & 59.32 & --\\
$+$AttnSR & \textbf{63.99} & \textbf{+4.67}\\
\midrule
\multicolumn{3}{l}{\textit{Qwen3-4B-Instruct-2507}}\\
PPO & 87.96 & --\\
$+$AttnSR & \textbf{91.24} & \textbf{+3.28}\\
\midrule
\multicolumn{3}{l}{\textit{Qwen2.5-3B-Instruct}}\\
PPO & 79.56 & --\\
$+$AttnSR & \textbf{87.23} & \textbf{+7.66}\\
\addlinespace
GRPO & 12.41 & --\\
$+$AttnSR & \textbf{20.07} & \textbf{+7.66}\\
\bottomrule
\end{tabular}

\end{minipage}\hfill
\begin{minipage}[t]{0.55\linewidth}
\caption{Full-set trajectory statistics. Steps count environment interactions, and Out. is output in thousands of characters, including observations.}
\label{tab:trajectory-overall}
\centering
\scriptsize
\setlength{\tabcolsep}{3pt}
\begin{tabular}{llrrrr}
\toprule
& & \multicolumn{3}{c}{Environment steps} & Out.\\
\cmidrule(lr){3-5}
Task & Method & Mean & Med. & P90 & (k)\\
\midrule
ALFWorld & GRPO$^{\dagger}$ & 36.93 & 11.0 & 105.0 & 33.44\\
 & PPO & 19.28 & 9.0 & 38.7 & 16.91\\
 & +AttnSR & \textbf{15.51} & \textbf{8.0} & \textbf{31.0} & \textbf{13.32}\\
\midrule
WebShop & GRPO$^{\dagger}$ & 8.46 & 6.0 & 12.0 & 11.65\\
 & PPO & \textbf{5.93} & \textbf{5.0} & \textbf{9.0} & \textbf{8.31}\\
 & +AttnSR & 8.03 & 7.0 & 12.0 & 11.08\\
\midrule
ScienceWorld & GRPO$^{\dagger}$ & 27.19 & 13.0 & \textbf{74.0} & 43.93\\
 & PPO & 30.68 & 15.0 & 80.0 & 45.42\\
 & +AttnSR & \textbf{21.51} & \textbf{10.0} & \textbf{74.0} & \textbf{31.99}\\
\bottomrule
\end{tabular}
\par\smallskip
\scriptsize $^{\dagger}$Same GRPO reference and checkpoint budgets as Table~\ref{tab:method-comparison}.

\end{minipage}
\end{table}

Table~\ref{tab:linear-checkpoints} in Appendix~\ref{app:full} shows that Linear Step Reward leads at epoch 1 but trails both methods at epochs 2--3 and \method{} at epoch 4. Joint changes to allocation, gating, and realized budgets prevent isolating attention contribution or ruling out other dense rewards. Differences are descriptive because per-task outputs were not retained for paired testing.

\paragraph{Supplementary Training Configuration.} Single-trajectory asynchronous Qwen2.5-7B-Instruct training gains 4.67 points in Table~\ref{tab:linear}. Changed collection, importance weighting, and critic updates preclude component attribution or direct comparison with canonical PPO, as discussed in Appendix~\ref{app:supplementary}.

\paragraph{Cross-backbone Check.} Qwen3-4B-Instruct-2507 gains 3.28 points, and Qwen2.5-3B-Instruct gains 7.66 points, rising from 218 to 239 successes in Table~\ref{tab:qwen25-3b}. As shown in Figure~\ref{fig:qwen25-3b-checkpoints} in the appendix, the 3B model improves on both seen and unseen tasks at all four checkpoints. Both retain the main PPO and shaping hyperparameters, and the 3B model uses relative-depth attention layers.

\paragraph{Extension to GRPO.} GRPO uses four rollouts per prompt for one epoch, matching the rollout count of four-epoch PPO with one rollout per prompt, but with fewer policy updates. 3B GRPO reweights group-normalized outcome advantages across action tokens, preserving each trajectory's mean advantage with $\alpha=0.25$, as detailed in Appendix~\ref{app:3b-grpo}. In Table~\ref{tab:qwen25-3b}, success rises from 34 to 55 tasks. Format-error rates of 36.86\% and 34.67\% and single runs limit interpretation.

\subsection{Learning Dynamics and Task-level Analysis}
\label{sec:learning-analysis}
\begin{figure}[!t]
\centering
\includegraphics[width=0.85\linewidth]{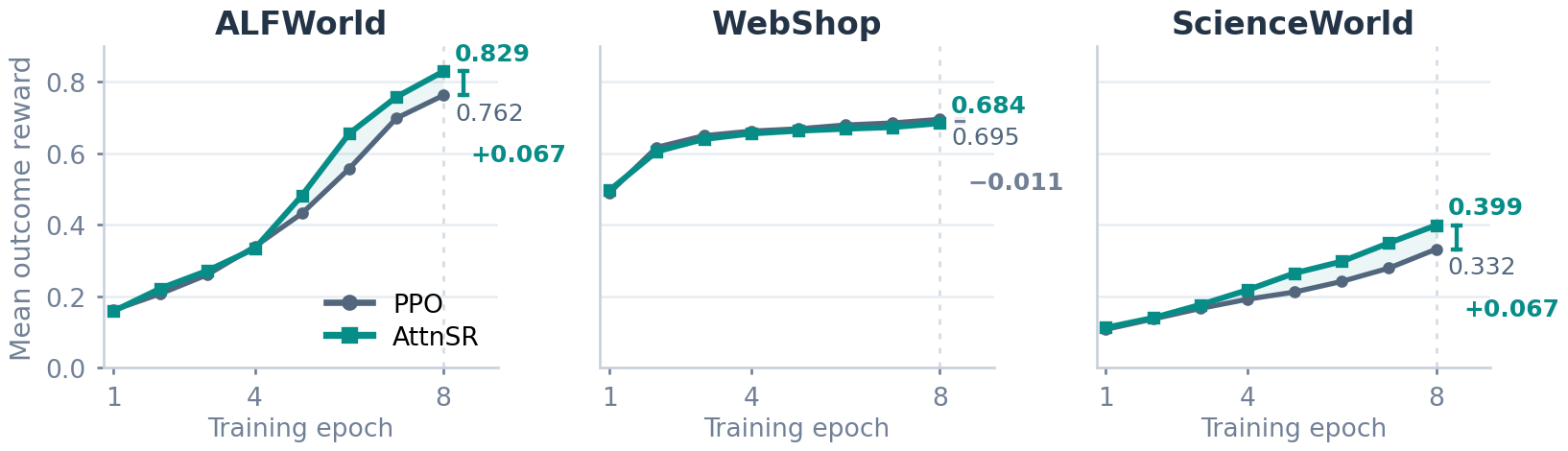}
\caption{Environment outcome reward during eight-epoch training (learning rate 1e-6), excluding shaping rewards. These traces supplement the four-epoch held-out comparison.}
\label{fig:training}
\end{figure}

\paragraph{Training and Checkpoints.} As reported in Appendix~\ref{app:full}, \method{} scores higher at all four ALFWorld and ScienceWorld checkpoints, with the largest gaps at epochs 4 and 3, respectively. WebShop gains at epochs 1--3 reverse slightly at epoch 4. The eight-epoch traces with learning rate 1e-6 in Figure~\ref{fig:training} supplement these results, but neither establishes an asymptotic advantage.

\paragraph{Generalization and Task Categories.} ALFWorld success rises from 92.86\% to 96.43\% on seen tasks and 89.55\% to 97.76\% on unseen tasks, as shown in Table~\ref{tab:method-comparison}. \emph{Clean}, \emph{Heat}, \emph{Cool}, and \emph{Pick2} gain 6.89--10.87 points, while \emph{Pick} and \emph{Look} remain unchanged, as listed in Table~\ref{tab:alfworld-category} in the appendix. Gains on object-processing and two-object tasks are consistent with benefits on dependent actions, but amount to only three to five successes per family. PPO already reaches 100\%/96.77\% on Pick/Look, so complexity is confounded with headroom.

\subsection{Inference-time Interaction Analysis}
\label{sec:interaction-analysis}
Table~\ref{tab:trajectory-overall} covers 5,186 PPO/AttnSR and 2,593 GRPO trajectories. Steps and output length do not measure reasoning or generated-token compute. The paired analyses in Appendix~\ref{app:trajectory-diagnostics} cover only the PPO and AttnSR policies.

\paragraph{Overall Interaction Cost.} Mean steps fall from 19.28 to 15.51 on ALFWorld ($-19.56\%$) and from 30.68 to 21.51 on ScienceWorld ($-29.88\%$), with higher success. Serialized output falls by 21.22\% and 29.56\%, respectively. WebShop rises from 5.93 to 8.03 steps without a clear gain.

\paragraph{What Explains Shorter Trajectories?} On 268 ScienceWorld tasks solved by both policies, mean steps fall from 11.06 to 7.40. The 246 joint ALFWorld successes instead increase from 11.19 to 11.99 steps, and the 157 joint WebShop successes from 5.11 to 6.19, as shown in Table~\ref{tab:trajectory-paired} in the appendix. The aggregate ALFWorld saving accompanies fewer prolonged failures, since the 24 PPO failures average 103.83 steps versus 109.12 for the 8 AttnSR failures. As discussed in Appendix~\ref{app:trajectory-diagnostics}, earlier unsuccessful stopping may also reduce aggregate length. These outcome-selected comparisons distinguish shorter solutions from fewer failures but do not establish a causal mechanism.

\paragraph{Scope of the Benefit.} WebShop has shorter PPO trajectories (5.93 steps versus 19.28/30.68 on ALFWorld/ScienceWorld), relatively close PPO/GRPO scores (70.74\%/67.21\%), and no detectable AttnSR gain. This suggests less benefit when shorter trajectories reduce delayed-credit difficulties. However, as noted in Table~\ref{tab:method-comparison}, PPO and GRPO optimization budgets differ, so this comparison establishes neither equivalence nor convergence. Policy-dependent lengths also confound horizon and complexity. Greater benefit on complex, long-horizon tasks remains a hypothesis, not evidence that step rewards generally fail on short tasks.

\section{Related Work}
\paragraph{Agentic RL.} Language agents interleave reasoning with actions~\citep{yao2023react,putta2024agentq}, each spanning multiple tokens. This creates a distinction between token-level optimization and step-level credit assignment. POAD decomposes Bellman backups within and across actions~\citep{wen2024poad}, ArCHer pairs a turn-level critic with a token-level actor~\citep{zhou2024archer}, and GiGPO groups steps with matching anchor states to estimate critic-free relative advantages~\citep{feng2025gigpo}. These methods address the granularity of value or advantage estimation. \method{} instead intervenes at the reward interface. It assigns outcome-conditioned rewards to interaction steps, which then enter the target computation of RL algorithms. In our token-level PPO implementation, these rewards are placed at assistant-step endpoints and propagated through GAE~\citep{schulman2017ppo,schulman2016gae}, retaining the critic architecture and optimization losses. We evaluate this reward interface in PPO and separately test attention-based token-advantage reweighting in GRPO.

\paragraph{Credit assignment.} Long horizons and partial observability complicate credit assignment from terminal outcomes~\citep{zhang2026creditsurvey}. VinePPO estimates intermediate values through Monte Carlo rollouts~\citep{kazemnejad2025vineppo}, AgentPRM learns step-level promise and progress~\citep{xi2025agentprm}, TARL uses an LLM judge for turn-level process supervision~\citep{tan2025tarl}, and SWEET-RL trains an evaluator with privileged information~\citep{zhou2025sweetrl}. Process supervision also uses step annotations~\citep{lightman2023verify}, estimated progress~\citep{setlur2024rewarding}, or trajectory preferences~\citep{liu2025oprl}. Other approaches use counterfactual baselines~\citep{foerster2018coma}, hindsight estimators~\citep{tan2026hcapo}, learned progress in SPA-RL~\citep{wang2025sparl}, or gradient saliency in ACPO~\citep{yin2025acpo}. FlowTracer derives token credit from answer-targeted information flow on an attention-induced graph~\citep{dong2026flowtracer}. \method{} draws its allocation prior from actor attention, avoiding additional environment rollouts, a separately trained large reward model, or an attribution backward pass. This signal expresses relevance rather than verified causal importance~\citep{jain2019attention,abnar2020quantifying}. Confidence gating moderates its strength, while the centered outcome determines its sign. Unlike return-preserving redistribution in RUDDER~\citep{arjona-medina2019rudder}, \method{} retains the outcome reward and adds intermediate rewards. It thus changes the return without guaranteeing policy invariance, and we evaluate on the original metrics.

\section{Conclusion}
\method{} allocates outcome-conditioned rewards using actor attention, without extra rollouts or a learned reward model. With PPO, ALFWorld success rises from 91.2\% to 97.1\%, exceeding the linear control. ScienceWorld shows numerical gains with fewer interactions, while WebShop has no statistically detectable gain. Supplementary experiments show gains under both PPO and a GRPO token-advantage reweighting variant, with one training run per method. Future work will extend the evaluation to longer context lengths and interaction horizons, examining how attention-based credit allocation handles action dependencies spanning more distant steps.

\subsection*{Ethics Statement}
This work studies training methods for simulated text-based agents. The experiments use benchmark environments and do not involve human subjects or collection of personal data. As with other agent-training methods, improved planning and tool-use capability could be transferred to applications with unintended consequences. Deployment should thus include task-specific safeguards and monitoring.

\subsection*{Data Availability Statement}
We use publicly available ALFWorld, WebShop, and ScienceWorld benchmark environments and do not introduce a new dataset. The evaluation splits, checkpoint-level summaries, and derived tables used in this paper are included with the supplementary artifacts.

\subsection*{Reproducibility Statement}
The training configuration, reward equations, evaluation splits, and checkpoint-level results are specified in this paper and its appendix. An anonymized repository will provide the reward extraction and allocation implementation, launch configurations, evaluation summaries, experiment record, significance analysis, and cached per-episode inputs. The main experiments use Qwen2.5-7B-Instruct, LoRA rank 64, VERL/Megatron training, vLLM rollout, and the fixed held-out sets of Section~\ref{sec:experimental-setup}.

\subsection*{AI Use Statement}
Generative AI tools were used to assist with the organization and language editing of this manuscript, as well as with the development, debugging, and refinement of experimental code. The authors reviewed and verified all AI-assisted content, including technical claims, equations, experimental results, implementation details, and code, against the repository artifacts and experimental records. The authors remain fully responsible for the correctness, integrity, and reproducibility of this work.

\bibliography{iclr2027_conference}
\bibliographystyle{iclr2027_conference}

\appendix
\section{Full checkpoint results}
\label{app:full}
\begin{table}[h]
\caption{Official scores (\%) at four checkpoints from the canonical held-out evaluations. ALFWorld official score equals binary success.}
\begin{center}
\small
\begin{tabular}{llrrrr}
\toprule
Task & Metric & Epoch 1 & Epoch 2 & Epoch 3 & Epoch 4\\
\midrule
\alf{} PPO & score (\%) & 36.50 & 69.34 & 89.42 & 91.24\\
\alf{} \method{} & score (\%) & 37.59 & 70.44 & 91.24 & 97.08\\
\webshop{} PPO & score (\%) & 63.72 & 67.53 & 66.71 & 70.74\\
\webshop{} \method{} & score (\%) & 65.74 & 68.35 & 69.58 & 69.71\\
\science{} PPO & score (\%) & 22.89 & 31.49 & 34.59 & 44.53\\
\science{} \method{} & score (\%) & 27.12 & 35.57 & 41.15 & 46.35\\
\bottomrule
\end{tabular}
\end{center}
\end{table}

\begin{table}[htbp]
\caption{ALFWorld success (\%) across checkpoints for the canonical positional control. All checkpoints use the full 274-task held-out set.}
\label{tab:linear-checkpoints}
\centering
\small
\begin{tabular}{lrrrr}
\toprule
Method & Epoch 1 & Epoch 2 & Epoch 3 & Epoch 4\\
\midrule
PPO & 36.50 & 69.34 & 89.42 & 91.24\\
Linear Step Reward & 43.43 & 59.85 & 79.20 & 92.34\\
AttnSR & 37.59 & 70.44 & 91.24 & 97.08\\
\bottomrule
\end{tabular}
\end{table}

\section{ALFWorld category-level results}
\label{app:alfworld-category}
\begin{table}[htbp]
\caption{Fine-grained ALFWorld task-family success at epoch 4 on the full held-out set. Columns use the compact family names \emph{Pick} (pick and place simple), \emph{Look} (look at object in light), \emph{Clean}, \emph{Heat}, \emph{Cool}, and \emph{Pick2} (pick two objects and place). Counts sum to 274 tasks.}
\label{tab:alfworld-category}
\centering
\scriptsize
\setlength{\tabcolsep}{5pt}
\begin{tabular}{lrrrrrrr}
\toprule
 & Pick & Look & Clean & Heat & Cool & Pick2 & Overall\\
\midrule
n & 59 & 31 & 58 & 39 & 46 & 41 & 274\\
PPO (\%) & 100.00 & 96.77 & 86.21 & 89.74 & 86.96 & 87.80 & 91.24\\
AttnSR (\%) & 100.00 & 96.77 & 93.10 & 100.00 & 97.83 & 95.12 & 97.08\\
$\Delta$ (pp) & 0.00 & 0.00 & +6.89 & +10.26 & +10.87 & +7.32 & +5.84\\
\bottomrule
\end{tabular}
\end{table}

\section{Implementation details}
\label{app:implementation}
The selected layers are 8, 10, 12, 14, 16, and 18. We use full causal-prefix softmax attention and do not renormalize within the step submatrix. AA mass targets historical assistant tokens, while AO mass targets the observation paired with the historical step. Step rewards are placed at assistant-step end tokens, and tool observations are excluded from the actor loss. The main implementation uses query blocks of 128 tokens and reconstructs tensor/context-parallel shards before graph aggregation. Format errors are assigned the ordinary outcome reward of zero rather than an extra penalty.

Each benchmark is trained independently from the instruction-tuned checkpoint. Table~\ref{tab:training-size} reports the training manifests and collected-batch updates. Epoch lengths use complete batches of eight, dropping the incomplete final batch when present. Training reward is binary success on ALFWorld, the official score in $[0,1]$ on WebShop, and the final ScienceWorld score clipped to $[0,100]$ and divided by 100. The native ScienceWorld step-reward increments are not supplied to PPO. Benchmark-specific expert-action and search bonuses are disabled.

\begin{table}[htbp]
\caption{Canonical training-set sizes and update budgets, per method. One update uses one collected batch of eight trajectories.}
\label{tab:training-size}
\centering
\small
\begin{tabular}{lrrr}
\toprule
Benchmark & Training tasks & Updates/epoch & Updates at epoch 4\\
\midrule
ALFWorld & 3,553 & 444 & 1,776\\
WebShop & 10,587 & 1,323 & 5,292\\
ScienceWorld & 3,592 & 449 & 1,796\\
\bottomrule
\end{tabular}
\end{table}

\begin{table}[h]
\caption{Framework and model configuration for the canonical runs.}
\label{tab:app-framework}
\centering
\scriptsize
\setlength{\tabcolsep}{5pt}
\begin{tabular}{ll}
\toprule
Component & Configuration\\
\midrule
Base model & Qwen2.5-7B-Instruct\\
Training framework & VeRL with Megatron-based training\\
Rollout engine & vLLM\\
Hardware & 8$\times$H200 GPUs; 140\,GB memory per GPU\\
Fine-tuning & LoRA rank 64; frozen base model\\
Trainable modules & Actor and critic LoRA adapters\\
Message format & Qwen standard chat format with JSON tool calls\\
Maximum context & 32,768 tokens\\
Additional length limits & No per-turn token cap or turn-count cap\\
Training parallelism & Megatron TP=4, CP=2\\
Rollout parallelism & Tensor parallelism 4\\
\bottomrule
\end{tabular}
\end{table}

\begin{table}[h]
\caption{PPO, evaluation, and AttnSR hyperparameters used for the main results.}
\label{tab:app-hparams}
\centering
\scriptsize
\setlength{\tabcolsep}{5pt}
\begin{tabular}{ll}
\toprule
Component & Configuration\\
\midrule
Batch size & 8\\
PPO update epochs & 1\\
Rollouts per prompt & 1\\
Advantage estimator & GAE with trajectory-level agent advantage\\
Actor / critic learning rate & $5\times10^{-6}$ / $5\times10^{-6}$\\
Reference-policy KL coefficient & 0.02\\
Checkpoint frequency & Once per data epoch\\
Decoding & Greedy; one rollout per held-out task\\
Evaluation sets & ALFWorld 274; WebShop 500; ScienceWorld 1,819\\
Attention layers & 8, 10, 12, 14, 16, 18\\
Attention aggregation & Full causal-prefix softmax; no step-submatrix renormalization; AA/AO = 0.5/0.5\\
Query block size & 128 tokens\\
Step-reward budget scale & 0.25\\
Global reward mean & First-batch mean initialization; stop-gradient EMA coefficient 0.99\\
Hop decay & 0.85\\
Temporal-bias correction & Weight $-0.25$; distance scale $\kappa=4$\\
Target attention mass / focus & 0.20 / 0.15\\
Reward placement & Final token of each assistant step; outcome reward on final step\\
\bottomrule
\end{tabular}
\end{table}

\section{Paired significance analysis}
\label{app:significance}
Table~\ref{tab:significance} reports the full uncertainty analysis summarized in Section~\ref{sec:results}. Both policies decode greedily on identical held-out tasks, so the final-checkpoint comparison is paired at the task level over the exact-input-matched episodes defined in Appendix~\ref{app:trajectory-diagnostics}. Wins and losses count tasks solved by exactly one policy, and the McNemar test is applied to those discordant pairs only. Bootstrap intervals resample the paired per-task differences 10,000 times.

\begin{table}[t]
\caption{Paired evaluation-set uncertainty for the final-checkpoint comparison, computed over exact-input-matched episodes. Wins and losses count tasks solved by exactly one policy, and $p$ is the unadjusted two-sided exact McNemar test on those discordant pairs. Intervals are 10,000-resample percentile bootstrap intervals for the mean paired difference (AttnSR $-$ PPO) in points. ALFWorld official score equals binary success. Intervals quantify sampling uncertainty over matched tasks.}
\label{tab:significance}
\centering
\small
\setlength{\tabcolsep}{4pt}
\begin{tabular}{lrrrrcc}
\toprule
Task & Pairs & Wins & Losses & McNemar $p$ & $\Delta$ Success (pp) & $\Delta$ Score (pp)\\
\midrule
ALFWorld & 274 & 20 & 4 & 0.002 & +5.84 [+2.55, +9.49] & +5.84 [+2.55, +9.49]\\
WebShop & 500 & 48 & 44 & 0.755 & +0.80 [-3.00, +4.40] & -1.04 [-3.36, +1.20]\\
ScienceWorld & 1488 & 173 & 136 & 0.040 & +2.49 [+0.20, +4.70] & +1.85 [-0.37, +4.06]\\
\bottomrule
\end{tabular}
\end{table}

\section{Inference-trajectory diagnostics}
\label{app:trajectory-diagnostics}
\paragraph{Source and measurement.}
We re-read the six final-checkpoint \texttt{generations/*\_openai.jsonl} files referenced by the canonical evaluation CSVs. The analysis includes 274 ALFWorld, 500 WebShop, and 1,819 ScienceWorld episodes per method. Recomputed means for official score, success, environment steps, parsed calls, serialized output length, and trace-parser events agree with the evaluation summaries. Per-episode numeric extracts and source-file hashes are retained with the analysis script, with no new training or evaluation.

Table~\ref{tab:trajectory-overall} additionally reports the three historical one-epoch GRPO evaluations on the same full-set sizes (2,593 trajectories). We extract their environment steps and serialized output lengths using the same definitions, and verify score, success, format-error rate, natural termination, and mean steps against the original GRPO summaries. Source hashes and numeric extracts are retained separately. These GRPO statistics are excluded from the paired and significance analyses below.

Environment steps use the evaluator \texttt{env\_steps} field. Tool calls are counted from \texttt{openai\_tool\_calls}. They are parseable calls, not necessarily valid or successful environment actions. In these exports, parsed assistant tool-call turns equal the call count and are not reported as an independent measure of reasoning depth. Output length counts characters in the serialized \texttt{output}, including tool-call syntax and observations. We do not interpret it as model-generated token count, internal reasoning length, or inference FLOPs. Medians and P90 values are computed from the episode-level observations, and P90 uses linear interpolation.

\paragraph{Outcome-conditioned and paired comparisons.}
Table~\ref{tab:trajectory-outcomes} separates successful and unsuccessful episodes using the evaluator binary success flag. These two groups have different task membership under the two policies. The paired analysis in Table~\ref{tab:trajectory-paired} additionally requires the same evaluation row, an exact input-text hash match, and an input unique within each benchmark set. This retains all 274 ALFWorld and 500 WebShop pairs. For ScienceWorld, 331 episodes are excluded because inputs differ or are non-unique, leaving 1,488 eligible pairs and 268 joint successes. No matching is inferred from row position alone.

\begin{table}[t]
\caption{Full-set trajectory lengths on each benchmark. Success and failure groups differ across policies and are not matched task subsets. Table~\ref{tab:trajectory-paired} provides the restricted paired comparison.}
\label{tab:trajectory-outcomes}
\centering
\small
\setlength{\tabcolsep}{4pt}
\begin{tabular}{llrrrr}
\toprule
& & \multicolumn{2}{c}{Successful} & \multicolumn{2}{c}{Unsuccessful}\\
Task & Method & $n$ & Mean steps & $n$ & Mean steps\\
\midrule
ALFWorld & PPO & 250 & 11.16 & 24 & 103.83\\
 & AttnSR & 266 & 12.69 & 8 & 109.12\\
\addlinespace
WebShop & PPO & 201 & 5.22 & 299 & 6.41\\
 & AttnSR & 205 & 6.62 & 295 & 9.00\\
\addlinespace
ScienceWorld & PPO & 520 & 11.32 & 1299 & 38.43\\
 & AttnSR & 562 & 8.00 & 1257 & 27.55\\
\bottomrule
\end{tabular}
\end{table}

\begin{table}[t]
\caption{Environment steps on exact-input-matched episodes where both policies succeed. Eligible pairs require the same evaluation row and identical input text, unique within each benchmark set. All ALFWorld and WebShop episodes qualify, while ScienceWorld retains 1,488 of 1,819 episodes, excluding 331 with different or non-unique input text. Results are descriptive.}
\label{tab:trajectory-paired}
\centering
\small
\setlength{\tabcolsep}{4pt}
\begin{tabular}{lrrrrr}
\toprule
Task & Eligible & Both succeed & PPO & AttnSR & $\Delta$ steps\\
\midrule
ALFWorld & 274 & 246 & 11.19 & 11.99 & +0.80\\
WebShop & 500 & 157 & 5.11 & 6.19 & +1.08\\
ScienceWorld & 1488 & 268 & 11.06 & 7.40 & -3.66\\
\bottomrule
\end{tabular}
\end{table}

On ALFWorld, the verified pairs include 20 tasks solved only by AttnSR, 4 only by PPO, 246 by both, and 4 by neither. This yields the observed net gain of 16 successes without claiming that every previously solved task remains solved. The corresponding WebShop counts are 48, 44, 157, and 251. Within the restricted ScienceWorld eligible set, they are 173, 136, 268, and 911. These counts should not replace the full-test success totals. Successful-subset comparisons are descriptive and post-selected on outcomes of both policies, not estimates of a causal treatment effect.

\paragraph{Tool use and termination.}
Table~\ref{tab:trajectory-errors} distinguishes exported trace-parser events from the evaluator terminal format-error flag. Most parser events are unmatched call/response pairs, not malformed JSON. Specifically, ALFWorld PPO has 23 missing-response events, while AttnSR has 8 missing-response and 4 orphan-response events. ScienceWorld PPO has 426 missing-response and 10 call-parsing events, while AttnSR has 245 missing-response events and no call-parsing events. WebShop has zero events under PPO and 3 missing-response events under AttnSR. An episode may contain multiple events, so the event count per episode differs from the fraction of affected episodes. We do not equate these exported trace diagnostics with runtime tool-execution failures.

\begin{table}[t]
\caption{Tool-trace and termination diagnostics at epoch 4. Parse/ep. counts exported trace-parser events, and Parse\% is the fraction of episodes with any such event. Format\% and Natural\% are the evaluator terminal flags. Repeat\% is the pooled fraction of adjacent parsed-call pairs with identical function names and arguments, and repetition need not imply an invalid or redundant action.}
\label{tab:trajectory-errors}
\centering
\small
\setlength{\tabcolsep}{4pt}
\begin{tabular}{llrrrrr}
\toprule
Task & Method & Parse/ep. & Parse\% & Format\% & Natural\% & Repeat\%\\
\midrule
ALFWorld & PPO & 0.084 & 8.39 & 0.00 & 0.00 & 0.04\\
 & AttnSR & 0.044 & 3.28 & 0.00 & 0.00 & 0.00\\
\addlinespace
WebShop & PPO & 0.000 & 0.00 & 0.00 & 0.00 & 0.04\\
 & AttnSR & 0.006 & 0.60 & 0.00 & 0.00 & 0.03\\
\addlinespace
ScienceWorld & PPO & 0.240 & 23.97 & 0.60 & 12.15 & 15.35\\
 & AttnSR & 0.135 & 13.47 & 0.22 & 18.42 & 16.30\\
\bottomrule
\end{tabular}
\end{table}

Natural termination is also not a success indicator. On ScienceWorld its rate increases from 12.15\% to 18.42\%, so earlier unsuccessful stopping may contribute to the full-set length reduction, which is why we separately report jointly successful trajectories. Exact adjacent-call repetition does not uniformly decrease either, as the pooled ScienceWorld repetition fraction increases from 15.35\% to 16.30\%. Repeated commands can be necessary in stateful environments and are not labeled invalid or redundant without inspecting the intervening state. On WebShop, the parsed-call totals are 824 search and 2,143 click calls for PPO, versus 1,088 search and 2,925 click calls for AttnSR. This confirms that the higher interaction count involves both search and navigation rather than one tool alone.

\section{Scope of supplementary runs}
\label{app:supplementary}
Table~\ref{tab:method-comparison} includes a one-epoch GRPO reference with four rollouts per prompt. Table~\ref{tab:linear} reports the canonical positional control, asynchronous and Qwen3 backbone comparisons, and the Qwen2.5-3B PPO and GRPO comparisons. The supplementary results are interpreted within their respective settings and are not pooled with the primary comparison for component attribution.

\paragraph{Asynchronous configuration.} Both variants use Qwen2.5-7B-Instruct, rank-64 actor and critic LoRA, batch size 8, actor/critic learning rates of $5\times10^{-6}$, a 32,768-token context, and four data epochs on ALFWorld. This configuration uses asynchronous collection with a replay queue and rollout-policy token log-probabilities. Direct token importance sampling masks ratios outside $[0.7,6.0]$. Observation tokens are skipped in GAE, with actor $\lambda=1-1/(1.5n_i)$ for $n_i$ trainable assistant tokens and critic $\lambda=1$. Two critic passes are performed per actor pass, and critic LoRA is restricted to MLP projections, with a trainable value head. The critic is initialized from the instruction-tuned model without additional value pretraining. The AttnSR variant retains the canonical attention layers and shaping hyperparameters. Thus the comparison changes more than collection timing relative to canonical PPO.

\paragraph{Qwen3 configuration.} The paired launch configuration uses Qwen3-4B-Instruct-2507 with the same main hyperparameters as the canonical study, namely rank-64 LoRA, actor/critic learning rates of $5\times10^{-6}$, batch and PPO minibatch size 8, one rollout per prompt, one PPO epoch per batch, reference KL coefficient 0.02, and a 32,768-token context. Training uses TP=4 and CP=2, and rollout uses TP=4. AttnSR retains the selected layers and reward settings. Four ALFWorld data epochs give 1,776 updates. Greedy evaluation of the epoch-4 policies on all 274 held-out tasks yields 241 successes for PPO and 250 for AttnSR. This is supplementary evidence across backbones, not a test isolating model size or individual reward components.

\subsection{Qwen2.5-3B PPO checkpoints}
\label{app:3b-ppo}
Both policies start from Qwen2.5-3B-Instruct with a frozen base and rank-64 actor/critic LoRA. The main settings are retained, including actor/critic learning rates $5\times10^{-6}$, batch and minibatch size 8, one rollout per prompt, one PPO epoch per batch, KL coefficient 0.02, and a 32,768-token context. Four data epochs give 1,776 updates, with checkpoints every 444 updates. AttnSR retains the reward budget scale 0.25, confidence targets, EMA, and attention aggregation settings in Table~\ref{tab:app-hparams}. To preserve relative depth in the 36-layer model, each zero-based reference index $l\in\{8,10,12,14,16,18\}$ is mapped to $\lfloor 35l/27+0.5\rfloor$, yielding $\{10,13,16,18,21,23\}$.

\begin{figure}[!htbp]
\centering
\includegraphics[width=\linewidth]{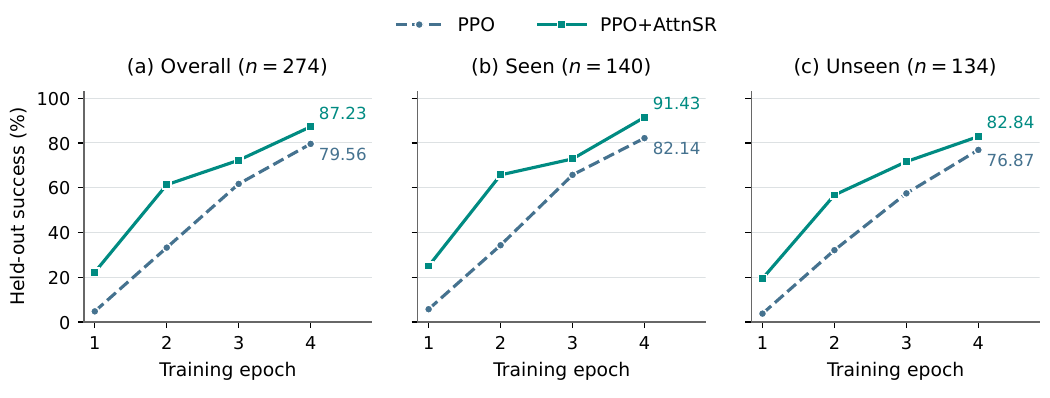}
\caption{Qwen2.5-3B-Instruct held-out ALFWorld success after each of four PPO training epochs, shown overall (274 tasks), seen (140), and unseen (134). Each point is a full-set greedy evaluation from one training run per method. Lines connect checkpoints without smoothing, and no estimate of training-seed uncertainty is available.}
\label{fig:qwen25-3b-checkpoints}
\end{figure}

Figure~\ref{fig:qwen25-3b-checkpoints} reports greedy evaluation of every checkpoint on all 274 held-out tasks. Overall success for PPO/AttnSR is 4.74/22.26\%, 33.21/61.31\%, 61.68/72.26\%, and 79.56/87.23\% at epochs 1--4. The final checkpoint is reported without best-checkpoint selection. At epoch 4, mean environment steps fall from 32.79 to 22.44, while format errors increase from 0.00\% to 2.55\%, so higher success does not imply uniform improvement in protocol compliance. Each method has one training run, so the curves do not estimate training-seed variability. The baseline and AttnSR source trees also contain critic/value-head compatibility differences beyond attention reward code, so these runs are a descriptive cross-backbone check rather than an isolated reward-component ablation.

\subsection{Qwen2.5-3B GRPO extension}
\label{app:3b-grpo}
The two GRPO policies independently start from Qwen2.5-3B-Instruct and train for one data epoch with four rollouts per prompt. Both use a frozen base, LoRA rank/alpha 64/64, actor learning rate $5\times10^{-6}$, prompt batch/minibatch size 8, actor KL loss coefficient 0.02, and no critic. Training uses TP=4, CP=2, rollout TP=4, and a 32,768-token context with prompt/response limits 7,168/25,600. The final checkpoint is saved at update 444. Evaluation uses greedy decoding on the same 274 tasks, with no training-time attention signal. Four rollouts for one epoch match the nominal trajectory count of one rollout for four PPO epochs, but not the update count, generated tokens, or runtime.

This experiment uses the initial \emph{token-advantage reweighting} AttnSR variant. Let $A_i$ be the ordinary GRPO outcome advantage computed from original environment rewards within each prompt group. Intermediate-step weights are broadcast to their action tokens and normalized to mean one over valid tokens with positive weights, accounting for step lengths. Set $\widetilde w_{i,u}=1$ at other valid action tokens, including the final step, unparsed positions, and zero-weight positions. The actor uses
\begin{equation}
A'_{i,u}=A_i\big[(1-\alpha)+\alpha\widetilde w_{i,u}\big],\qquad \alpha=0.25.
\label{eq:grpo-attnsr-reweight}
\end{equation}
Observations and padding are masked. This preserves each trajectory's mean advantage and its sign, while varying its magnitude across steps, and equal-reward groups retain zero advantage. The collector uses the same relative layers and attention allocation as the 3B PPO run. However, positive trajectory confidence is a common factor that cancels during token-weight normalization, and only zero confidence disables redistribution. Thus this variant does not retain PPO's continuous confidence modulation, outcome EMA, or intermediate rewards and GAE.

Both methods use the same GRPO implementation overlay, with attention reweighting enabled only in the AttnSR arm. Table~\ref{tab:qwen25-3b} shows final success of 34/274 for GRPO and 55/274 for GRPO+AttnSR. Seen success rises from 15.00\% to 22.14\% (21 to 31 of 140 tasks), and unseen success from 9.70\% to 17.91\% (13 to 24 of 134 tasks). Format-error counts remain high at 101/274 and 95/274. These single-run results are descriptive, with no repeated-seed or paired significance analysis.

\section{Additional ALFWorld baselines}
\label{app:additional-baselines}
Our ALFWorld adaptations of TARL~\citep{tan2025tarl}, AgentPRM~\citep{xi2025agentprm}, and FlowTracer~\citep{dong2026flowtracer} are trained independently from Qwen2.5-7B-Instruct with LoRA for four data epochs. We evaluate the final checkpoints at update 1,776 on all 274 held-out ALFWorld tasks (140 seen and 134 unseen), using one greedy rollout per task and a 32,768-token context. The implementation variants recorded in the experiment inventory are TARL-Turn, AgentPRM-Progress, and FlowTracer-PPO, respectively. Table~\ref{tab:method-comparison}(b) retains the canonical PPO and AttnSR results as references, and no paired significance comparison with the three additional methods is claimed.

Tables~\ref{tab:additional-baseline-outcomes} and~\ref{tab:additional-baseline-lengths} report the final-checkpoint details. Success counts are 265, 252, and 262 for TARL, AgentPRM, and FlowTracer. All three have zero terminal format errors, and AgentPRM has eight naturally terminated episodes. These diagnostic fields are distinct from task success and do not indicate whether training completed normally. The small number of failed tasks (9, 22, and 12) makes failure-conditioned step means sensitive to individual long trajectories. Category-level counts and outcome-conditioned lengths are descriptive, not causal evidence about credit assignment.

Text lengths are obtained by retokenizing saved evaluation text with \texttt{Qwen2TokenizerFast}, with \texttt{add\_special\_tokens=False}. Full trajectories include tool observations, while assistant-only counts exclude tool replies. These measurements do not recover the original sampled token IDs and should not be interpreted as exact generation costs. All results are from one training run per method, and no estimate of training-seed variability is available.

\begin{table}[!htbp]
\caption{Task outcomes of the additional ALFWorld baselines at the final checkpoint (update 1,776). (a)~Successful tasks, with success rates (\%) in parentheses. Format and termination columns count episodes. (b)~Successful tasks by category, with task counts in the header. Each row of (b) sums to the overall success count in (a).}
\label{tab:additional-baseline-outcomes}
\vspace{2pt}
\centering
\small
\renewcommand{\arraystretch}{1.1}
\setlength{\tabcolsep}{7pt}
\begin{tabular}{@{}lrrrrr@{}}
\multicolumn{6}{@{}l}{\textbf{(a)} Overall success and termination}\\
\toprule
Method & Overall / 274 & Seen / 140 & Unseen / 134 & Format errors & Natural term.\\
\midrule
TARL & 265 (96.72) & 135 (96.43) & 130 (97.01) & 0 & 0\\
AgentPRM & 252 (91.97) & 130 (92.86) & 122 (91.04) & 0 & 8\\
FlowTracer & 262 (95.62) & 135 (96.43) & 127 (94.78) & 0 & 0\\
\bottomrule
\end{tabular}

\vspace{10pt}
\begin{tabular}{@{}lrrrrrr@{}}
\multicolumn{7}{@{}l}{\textbf{(b)} Successful tasks by category}\\
\toprule
Method & Look (31) & Pick (59) & Clean (58) & Cool (46) & Heat (39) & Pick2 (41)\\
\midrule
TARL & 30 & 58 & 54 & 44 & 39 & 40\\
AgentPRM & 31 & 56 & 54 & 40 & 34 & 37\\
FlowTracer & 30 & 57 & 54 & 45 & 36 & 40\\
\bottomrule
\end{tabular}

\vspace{18pt}
\caption{Trajectory lengths of the additional ALFWorld baselines at the final checkpoint. (a)~Environment steps and mean serialized output length in thousands of characters. Successful and failed subsets differ across policies. (b)~Retokenized evaluation-text lengths. Full trajectories include tool observations, while assistant-only text excludes tool replies. These are counts from saved text, not original sampled token IDs or exact generation costs.}
\label{tab:additional-baseline-lengths}
\vspace{2pt}
\centering
\small
\renewcommand{\arraystretch}{1.1}
\setlength{\tabcolsep}{7pt}
\begin{tabular}{@{}lrrrrr@{}}
\multicolumn{6}{@{}l}{\textbf{(a)} Environment steps and output length}\\
\toprule
& \multicolumn{4}{c}{Environment steps} & Output\\
\cmidrule(lr){2-5}
Method & Mean & Median & Success mean & Failure mean & (k chars)\\
\midrule
TARL & 13.70 & 8 & 10.83 & 98.33 & 11.77\\
AgentPRM & 14.18 & 8 & 10.33 & 58.18 & 13.89\\
FlowTracer & 19.06 & 14 & 16.06 & 84.42 & 17.26\\
\bottomrule
\end{tabular}

\vspace{10pt}
\begin{tabular}{@{}lrrrrrr@{}}
\multicolumn{7}{@{}l}{\textbf{(b)} Retokenized text length}\\
\toprule
& \multicolumn{3}{c}{Full trajectory tokens} & \multicolumn{3}{c}{Assistant-only tokens}\\
\cmidrule(lr){2-4}\cmidrule(l){5-7}
Method & Mean & Median & P90 & Mean & Median & P90\\
\midrule
TARL & 3,983 & 2,656 & 7,824 & 348 & 214 & 641\\
AgentPRM & 4,596 & 2,673 & 9,493 & 648 & 306 & 1,303\\
FlowTracer & 5,837 & 4,203 & 10,186 & 480 & 348 & 768\\
\bottomrule
\end{tabular}
\end{table}

\section{Notation}
\label{app:notation}
Table~\ref{tab:notation} lists the symbols of Sections~\ref{sec:problem}--\ref{sec:method}. Episode index $i$ is omitted where unambiguous.

\begin{table}[h]
\caption{Notation. Hyperparameter values refer to the canonical AttnSR runs.}
\label{tab:notation}
\begin{center}
\small
\setlength{\tabcolsep}{4pt}
\begin{tabular}{ll}
\toprule
Symbol & Meaning \\
\midrule
$i$, $t$ & episode index; collected training-batch index \\
$K_i$ & number of assistant steps in episode $i$ \\
$k$, $j$ & step indices, $1\le k<j\le K_i$ \\
$S_{i,k}$ & interaction step: assistant message and observation, when present \\
$A_{i,k}$, $O_{i,k}$ & assistant message; associated environment observation \\
$\mathcal{I}_{i,k}$ & token indices in the assistant span of step $k$ \\
$u$ & token index into the serialized trajectory \\
$e_{i,k}$ & endpoint: last trainable token of step $k$ \\
$r_{i,u}$ & reward written at token $u$ \\
$R_i$, $\bar R_t$ & outcome reward; EMA updated using batch $t$ before shaping \\
$M_t$ & mean outcome reward in batch $t$; initializes $\bar R_1=M_1$ \\
$l$, $\mathcal{L}$, $L$ & layer index; selected layers; $L=|\mathcal{L}|=6$ \\
$G^{(l)}_{i,\mathrm{AA}}$, $G^{(l)}_{i,\mathrm{AO}}$ & step attention to historical assistant tokens; to paired observations \\
$G^{(l)}_i$ & mixed step attention, $0.5G^{(l)}_{i,\mathrm{AA}}+0.5G^{(l)}_{i,\mathrm{AO}}$ \\
$a^{(l)}_{i,k}$ & backtracked relevance of step $k$ at layer $l$, defined in Eq.~\ref{eq:backtrack} \\
$\Delta_{i,k}$, $d_{i,k}$ & steps from $k$ to the terminus, $K_i-k$; proximity feature $\exp(-(\Delta_{i,k}-1)/\kappa)$ \\
$p^{(l)}_{i,k}$, $p_{i,k}$ & per-layer allocation; geometric-consensus allocation, defined in Eq.~\ref{eq:consensus} \\
$m^{(l)}_i$, $H^{(l)}_i$ & terminal-step covered mass; normalized entropy of $p^{(l)}_i$ \\
$C_i$ & confidence gate, $C_i\in[0,1]$ \\
$B_i$, $s_{i,k}$ & signed shaping budget; step reward $B_ip_{i,k}$, defined in Eq.~\ref{eq:step-reward} \\
$\hat A_u$, $\widetilde A_u$, $y_u$ & unnormalized advantage; normalized actor advantage; critic target \\
$V^{\mathrm{old}}_u$, $V_\phi$ & pre-update critic value; critic network \\
$\operatorname{sg}$, $\operatorname{norm}$, $\operatorname{TV}$ & stop-gradient; standardization; total-variation distance \\
$\rho=0.85$ & hop decay in the backtracking recurrence \\
$\eta=-0.25$, $\kappa=4$ & recency-correction weight and proximity decay \\
$T=1$, $\epsilon=10^{-12}$ & allocation softmax temperature; lower bound before logarithms \\
$\beta=0.25$ & budget scale, bounding $|B_i|$ \\
\bottomrule
\end{tabular}
\end{center}
\end{table}

\end{document}